\documentclass[runningheads]{llncs}
\usepackage{graphicx}
\usepackage{tikz}
\usepackage{comment}
\usepackage{amsmath,amssymb} % define this before the line numbering.
\usepackage{color}
\usepackage{graphicx}
\usepackage{amsmath}
\usepackage{amssymb}
\usepackage{booktabs}
\usepackage{bm}
\usepackage{colortbl}
\usepackage{multirow}
\usepackage{graphics}
\usepackage{wrapfig}

\begin{document}
% \linenumbers
\pagestyle{headings}
\mainmatter
\def\ECCVSubNumber{3627}  % Insert your submission number here

\title{Convolutional Embedding Makes Hierarchical Vision Transformer Stronger}

%\begin{comment}
% \titlerunning{ECCV-22 submission ID \ECCVSubNumber} 
% \authorrunning{ECCV-22 submission ID \ECCVSubNumber} 
% \author{Anonymous ECCV submission}
% \institute{Paper ID \ECCVSubNumber}
%\end{comment}
%******************

    \makeatletter
\def\@fnsymbol#1{\ensuremath{\ifcase#1\or \dagger\or \ddagger\or
   \mathsection\or \mathparagraph\or \|\or **\or \dagger\dagger
   \or \ddagger\ddagger \else\@ctrerr\fi}}
    \makeatother

%\begin{comment}
\author{Cong Wang\inst{1,2} \and
Hongmin Xu 
\thanks{Corresponding author.}
\inst{1} \and
Xiong Zhang\inst{4} \and
Li Wang\inst{2} \and
Zhitong Zheng\inst{1} \and
Haifeng Liu\inst{1,3}}
\authorrunning{C. Wang et al.}
\institute{Data \& AI Engineering System, OPPO, Beijing, China\\
\email{\{wangcong3575, xhmjimmy\}@gmail.com, \{liam,blade\}@oppo.com}\\ \and
Beijing Key Lab of Urban Intelligent Traffic Control Technology, North China University of Technology, Beijing, China\\
\email{li.wang@ncut.edu.cn} \and
University of Science and Technology of China, Hefei, China\\ \and
Neolix Autonomous Vehicle, Beijing, China\\
\email{zhangxiong@neolix.cn}
}
%\end{comment}
%******************
\maketitle

\begin{abstract}
Vision Transformers (ViTs) have recently dominated a range of computer vision tasks, yet it suffers from low training data efficiency and inferior local semantic representation capability without appropriate inductive bias.
Convolutional neural networks (CNNs) inherently capture regional-aware semantics, inspiring researchers to introduce CNNs back into the architecture of the ViTs to provide desirable inductive bias for ViTs.
However, \textit{is the locality achieved by the micro-level CNNs embedded in ViTs good enough}? 
In this paper, we investigate the problem by profoundly exploring how the macro architecture of the hybrid CNNs/ViTs enhances the performances of hierarchical ViTs. 
Particularly, we study the role of token embedding layers, alias \textit{convolutional embedding} (CE), and systemically reveal how CE injects desirable inductive bias in ViTs. 
Besides, we apply the optimal CE configuration to $4$ recently released state-of-the-art ViTs, effectively boosting the corresponding performances.
Finally, a family of efficient hybrid CNNs/ViTs, dubbed \textbf{CETNet}s, are released, which may serve as generic vision backbones.
Specifically, CETNets achieve 84.9\% Top-1 accuracy on ImageNet-1K (training from scratch), 48.6\% box mAP on the COCO benchmark, and 51.6\% mIoU on the ADE20K, substantially improving the performances of the corresponding state-of-the-art baselines.

\keywords{Vision Transformers, Convolutional Neural Networks, Convolutional Embedding, Micro and Macro Design}
\end{abstract}

\section{Introduction}

% paragraph 1

% Why people are addicted into ViT?
% The advantages.

Over the last decades, convolutional neural networks (CNNs) significantly succeeded in the computer vision community due to their inherent properties, including the translation invariance, the locality attention, and the sharing weight design.
Those characteristics prove critical for many tasks, such as image recognition \cite{deng2009imagenet,he2016deep}, semantic image segmentation\cite{chen2017deeplab,zhang2021dcnas}, and object detection\cite{dai2016r,ren2015faster}.
At the same time, researchers take a very different way in the natural language processing (NLP) field. 
Since the seminal work \cite{vaswani2017attention} demonstrated the extraordinary capability of the transformer by employing a unified yet simple architecture to tackle the machine translation task, transformers have become the de-facto architectures to resolve NLP tasks \cite{radford2018improving,devlin2018bert}.

\begin{figure}
\centering 
\includegraphics[width=0.75\textwidth]{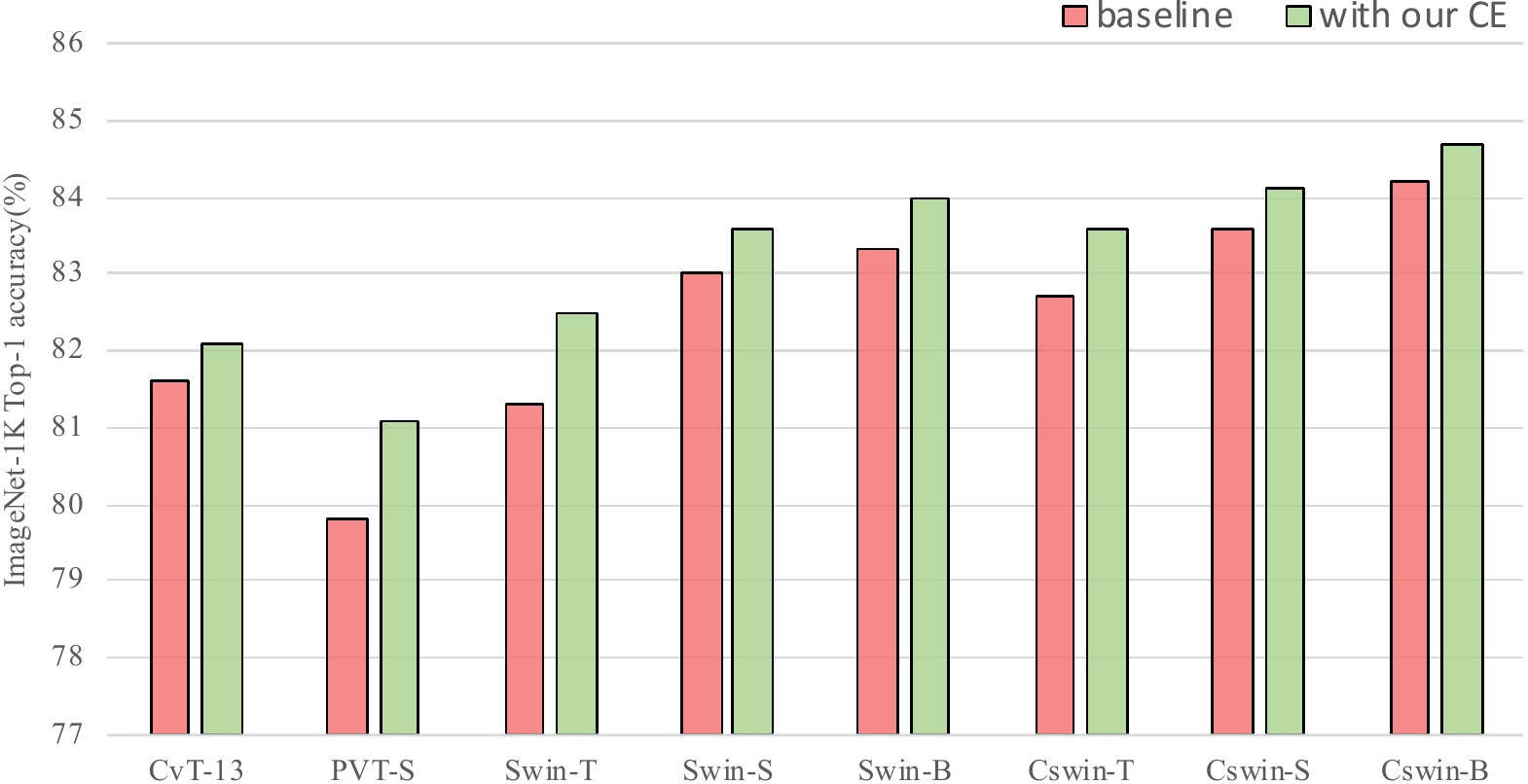} 
%\vspace{-10pt}
\caption{
\textbf{Performance Improvements of the Convolutional Embedding (CE).}
The figure presents the performance gains of the CE among 4 SOTA ViTs, \textit{i.e.,} \textit{CvT} \cite{wu2021cvt}, \textit{PVT} \cite{wang2021pyramid}, \textit{SWin}\cite{liu2021swin}, and \textit{CSWin}\cite{dong2021cswin} on the ImageNet-1K dataset, indicating that the CE indeed significantly improves the baseline methods
} 
\label{fig:teaser} 
\end{figure}

In the computer vision domain, certain works \cite{wang2018non,bello2019attention,srinivas2021bottleneck,hu2018squeeze} successfully brought the key idea of transformer, \textit{i.e.,} attention paradigm, into CNNs and achieved remarkable improvements. 
Naively transferring the transformer to image recognition, Dosovitskiy \textit{et al.}\cite{dosovitskiy2020image} demonstrated that the vanilla Vision Transformers (ViTs) structure could achieve comparable performance compared with the state-of-the-art (SOTA) approaches on the ImageNet-1K dataset \cite{dosovitskiy2020image}.
Further, pre-trained on the huge dataset JFT-300M \cite{sun2017revisiting}, the vanilla ViTs outperformed all the SOTA CNNs by a large margin and substantially advanced the SOTA, suggesting that ViTs may have a higher performance ceiling.

% Why right inductive bias is important?
% higher performance limitation. easier to train.
% paragraph 2
ViTs rely on highly flexible multi-head self-attention layers to favor dynamic attention, capture global semantics, and achieve good generalization ability. Yet, recent works find that lacking proper inductive bias, locality, ViTs are substandard optimizability \cite{xiao2021early}, low training sample-efficient \cite{d2021convit}, and poor to model the complex visual features and the local relation in an image \cite{marquardt2004token,wu2021cvt}. 
% Two main research line attempts to introducing locality. 
% How researchers introducing locality to ViT?
Most existing works attempt to introduce local mechanisms to ViTs in two paths. One line of work relives the inductive bias problem via non-convolutional ways. Liu \textit{et al.} \cite{liu2021swin,dong2021cswin,vaswani2021scaling} limits the attention computation in a local window to enable local receptive field for attention layers, making the overall network still maintain nearly pure attention-based architecture. 
Concurrently, as the CNNs are inherently efficient for their sliding-window manner, local receptive field, and inductive bias \cite{battaglia2018relational}, another line of work directly integrates CNNs into ViTs design to bring convolutional inductive bias into ViTs in a hard \cite{wu2021cvt,li2021localvit,yuan2021incorporating} or soft \cite{d2021convit,touvron2021training} way.
However, most of those works focus on modifying the micro design of ViTs to enable locality, which raises the question:

% What is the equivalant question?
% The overall goal is makes the whole network has the best performance. Thus, researcher do what to introduce the locality to VT.
% Which raise what question?
\textit{Is the inductive bias obtained via the micro design of ViTs good enough to empower the locality to ViTs?} or \textit{Can the macro architecture design of the network further introduce the desirable inductive bias to ViTs?}

% paragraph 3
% What findings makes us want to ask questions?
% Why we want to ask question?
We ask the questions mentioned above based on the following findings. The previous work, EarlyConv \cite{xiao2021early}, notices that simply replacing the original patchify stem with a 5-layer convolutional stem can yield 1-2\% top-1 accuracy on ImageNet-1K and improve the training stability of ViTs. Subsequently, CoAtNet \cite{dai2021coatnet} further explore the hybrid design of CNNs and ViTs based on the observation: the depth-wise convolution can be naturally integrated into the attention block. 
Meanwhile, those works suggest that the convolutional layers can efficiently introduce inductive bias in a shallow layer of the whole network. 
However, when we retrospect the roadmap of modern CNNs, at the late of 2012, after AlexNet \cite{krizhevsky2012imagenet} exhibited the potential of CNNs,  subsequent studies, for instance, VGG \cite{simonyan2014very}, ResNet \cite{he2016deep}, DenseNet \cite{huang2017densely}, EfficientNet \cite{tan2019efficientnet,tan2021efficientnetv2} ConvNet2020 \cite{liu2022convnet}, and \textit{etc.} reveal that: convolutions can represent the complex visual features even in a deep layer of network effectively and efficiently.
% Which means CNN can provide strong and effective prior of inductive bias.
Our research explores the macro network design with hybrid CNNs/ViTs. We want to bridge the gap between the pure CNNs network and the pure ViT network and extend the limitation of the hybrid CNNs/ViTs network.

% paragraph 4
% What is our hypothesis?
% to verify our hypothesis, what we do?? what is our motivation and experiments design.
To test this hypothesis, we start from CNNs' effective receptive field (ERF). As previous work of Luo \textit{et al.} \cite{luo2016understanding} points out, the output of CNNs considerably depends on their ERF. With larger ERF, CNNs can leave out no vital information, which leads to better prediction results or visual features. 
% The architecture designs such as subsampling, which has been successfully applied to hierarchical ViT \cite{liu2021swin, dong2021cswin}, is able to increase the ERF significantly. 
Under this perspective, our exploration is imposing strong and effective inductive bias for attention layers via the macro design of network architecture.
We specifically focus on \textit{patch embedding}, alias \textit{convolutional embedding} (CE), of the hierarchical ViTs architecture. 
CE locate at the beginning of each stage, as shown in Fig. \ref{fig:overall_architecture}. The CE aims to adjust the dimension and the number of tokens. Most following works also apply one or two convolutions embedding layers \cite{wu2021cvt,dong2021cswin,zhang2021rest,wang2021pyramid,li2021localvit}. 
However, these embedding layers cannot offer enough ERF to capture complex visual representations with desirable inductive bias. Since stacking more convolutional layers can increase the ERF \cite{luo2016understanding}, we construct a simple baseline with only 1-layer CE and gradually increase the number of convolutions layers in CE to get more variants. Meanwhile, keep changing FLOPs and parameter numbers as small as possible. We observe that the small change of CE in each stage results in a significant performance increase in the final model. 
% as shown in Table \ref{tab:Table_numberOfCE}.

% paragraph 5
% What we find? What the experiment results suggestion?
% What we do to understand the phenomenon (findings)?
% What conclusion we got?
% How we further use the conclusions?
% We people can do with our conclusions?

Based on extensive experiments, we further understand how CE affects the hybrid network design of CNNs/ViTs by injecting desirable inductive bias. We make several observations. 
% CE is important, but how?
1) CNNs bring strong inductive bias even in deep layers of a network, making the whole network easier to train and capturing much complex visual features. At the same time, ViTs allow the entire network has a higher generalization ceiling.
% 2) CE is able to impose effective inductive bias.  CE is able to empower most ViT-based backbone.
% Token-to-Token advocate this viewpoint either.
2) CE can impose effective inductive bias, yet different convolution layers show variable effectiveness. Besides, the large ERF is essential to designing CE or injecting the desirable inductive bias to ViTs, even though it is a traditional design in a pure CNNs network \cite{tan2021efficientnetv2,luo2016understanding}.
% 3) CNN can make transformer stronger in even deep of networks.
3) CNNs can help ViTs see better even in deep networks, providing valuable insights to guide how to design the hybrid CNNs/ViTs network.
% In deep of networks, CNN sometime can archive the same performanace in comprsion to ViT, if not better.
4) It is beneficial to combine the macro and micro introduce inductive bias to obtain a higher generalization ceiling of ViTs-based network.

% 3) visualization experiment reveals CNN make ViT extract even diverse attention map/feature map, which may hardly obtain via shallow convlution stacking or simple non-convolutional localty machanism.

Our results advocate the importance of CE and deep hybrid CNNs/ViTs design for vision tasks. ViT is a general version of CNN \cite{cordonnier2019relationship}, and tons of works have proven the high generalization of ViTs-based networks, which spurs researchers to line up to chase the performance ceiling with pure attention networks. 
After inductive bias is found to be crucial to significantly improve the training speed and sample-efficiency of ViTs, people's efforts are mainly devoted to creating the micro design of ViTs to enhance it \cite{liu2021swin,d2021convit}. 
Concurrently, EarlyConv  \cite{xiao2021early} and CoAtNet \cite{dai2021coatnet} verify the efficiency of convolutions in shallow layers of the ViT-based network. Our study further pushes the boundary of the macro design of the hybrid CNNs/ViTs network. 
Our results also suggest that even in deep layers of the ViTs network, correctly choosing the combination of CNNs/ViTs design, one can further improve the upper-performance limitation of the whole network. 
Finally, we propose a family of models of hybrid CNNs/ViTs as a generic vision backbone. 

% paragraph 6

To sum up, we hope our findings and discussions presented in this paper will deliver possible insights to the community and encourage people to rethink the value of CE in the hybrid CNNs/ViTs network design.

\section{Related Work}

\subsubsection{Convolutional neural networks.}
Since the breakthrough performance of AlexNet\cite{krizhevsky2012imagenet}, the computer vision field has been dominated by CNNs for many years. In the past decade, we have witnessed a steady stream of new ideas being proposed to make CNNs more effective and efficient\cite{simonyan2014very,szegedy2015going,he2016deep,huang2017densely,hu2018squeeze,howard2017mobilenets,sandler2018mobilenetv2,yu2015multi,tan2021efficientnetv2,radosavovic2020designing}. 
One line of work focuses on improving the individual convolutional layer except for the architectural advances. For example, the depthwise convolution \cite{xie2017aggregated} is widely used due to its lower computational cost and smaller parameter numbers, and the deformable convolution \cite{dai2017deformable} can adapt to shape, size, and other geometric deformations of different objects by adding displacement variables.
The dilated convolution \cite{yu2015multi} introduces a new parameter called "dilation rate"  into the convolution layer, which can arbitrarily expand the receptive field without additional parameters cost. 

%These CNNs architectures are still the primary backbones for computer vision tasks such as object detection\cite{ren2015faster}, instance segmentation\cite{he2017mask}, semantic segmentation\cite{long2015fully}, and so on.

\subsubsection{Vision Transformers.}
Transformer \cite{vaswani2017attention} has become a prevalent model architecture in natural language processing (NLP) \cite{devlin2018bert,brown2020language} for years. Inspired by the success of NLP, increasing effort on adapting Transformer to computer vision tasks. 
Dosovitskiy \textit{et al.} \cite{dosovitskiy2020image} is the pioneering work that proves pure Transformer-based architectures can attain very competitive results on image classification, shows strong potential of the Transformer architecture for handling computer vision tasks. 
The success of \cite{dosovitskiy2020image} further inspired the applications of Transformer to various vision tasks, such as image classification\cite{touvron2021training,yuan2021tokens,xu2021co,han2021transformer,wu2021cvt}, object detection\cite{carion2020end,zhu2020deformable,zheng2020end,dai2021up} and semantic segmentation\cite{wang2021max,strudel2021segmenter}. Furthermore, some recent works \cite{wang2021pyramid,zhang2021multi,liu2021swin} focusing on technique a general vision Transformer backbone for  general-purpose vision tasks. They all follow a hierarchical architecture and develop different self-attention mechanisms. The hierarchical design can produce multi-scale features that are beneficial for dense prediction tasks. An efficient self-attention mechanism reduced the computation complexity and enhanced modeling ability.

\subsubsection{Integrated CNNs and Transformer.}
CNNs are good at capturing local features and have the advantages of shift, scale, and distortion invariance, while Transformer have the properties of dynamic attention, global receptive field, and large model capacity. Combining convolutional and Transformer layers can achieve better model generalization and efficiency. Many researchers are trying to integrate CNNs and Transformer. Some methods\cite{wang2018non,bello2019attention,hu2019local,wu2020visual,srinivas2021bottleneck,shen2021efficient} attempt to augment CNNs backbones with self-attention modules or replace part of convolution blocks with Transformer layers. In
comparison, inspired by the success of ViT\cite{dosovitskiy2020image}, recent trials attempt to leverage some appropriate convolution properties to enhance Transformer backbones. ConViT\cite{d2021convit} introduce a parallel convolution branch to impose convolutional inductive biases into ViT\cite{dosovitskiy2020image}. Localvit \cite{li2021localvit} adding a depth-wise convolution in  Feed-Forward Networks(FFN) component to extract the locality, and CvT\cite{wu2021cvt} employs convolutional projection to calculate self-attention matrices to achieve additional modeling of local spatial context. Besides the “internal” fusion, some method\cite{dai2021coatnet,xiao2021early} focus on structural combinations of Transformer and CNNs.

\section{Hybrid CNNs/ViTs Network Design}

\begin{figure*}[ht] 
\centering 
\includegraphics[width=1\textwidth]{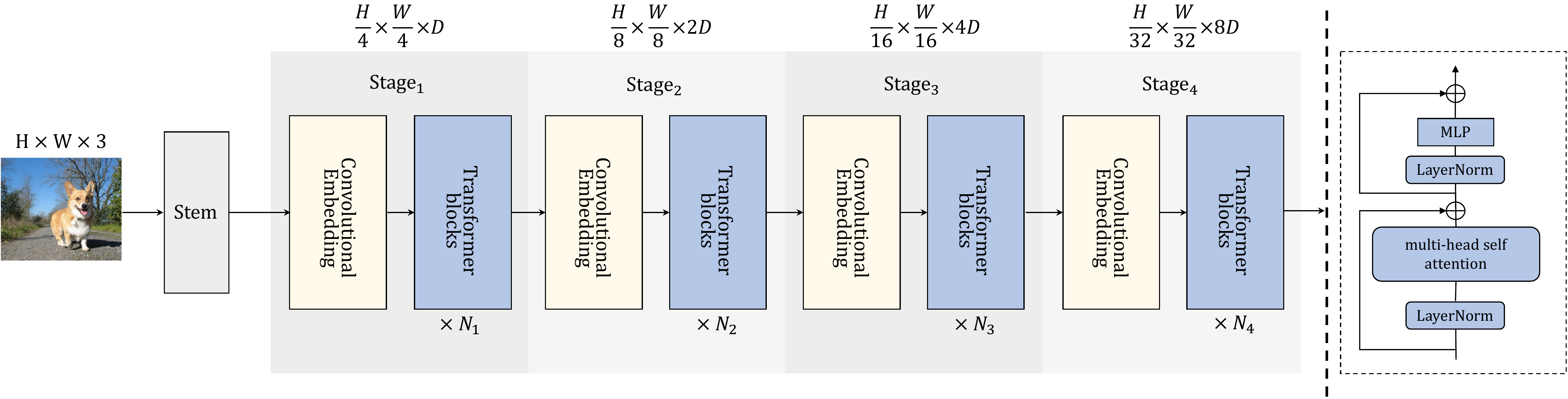} 
\caption{Left: the overall architecture of a general hierarchical ViTs network, which consists of two main parts: \textit{Stem} and \textit{Stages}. Each stage contains a \textit{convolutional embedding} and a set of \textit{ViTs blocks}. Right: an example of a standard \textit{ViTs block}\cite{dosovitskiy2020image}}
\label{fig:overall_architecture} 
\end{figure*}

A general hierarchical ViTs model architecture is illustrated in Figure \ref{fig:overall_architecture}. 
The convolutional stem is applied for an input image to extract the low-level feature, followed typically four stages to extract diverse-scale deep representations gradually. Each stage consists of a \textit{convolutional embedding} (CE) block and a set of ViTs blocks. 
More specifically, the CE block is located at the beginning of each stage and aims to adjust the dimension, change the number, and reduce the resolution of input tokens. 
For two adjacent stages, the reduction factor is set to 2. After that, those tokens are fed to the following ViTs blocks to generate a global context. To sum up, the proposed network contains five stages $(S0, S1, S2, S3, S4)$, where $S0$ is the convolutional stem.

The original hierarchical ViTs architecture \cite{liu2021swin} is following the traditional design of the CNNs network, VGG \cite{simonyan2014very}, targeting giving the tokens the ability to represent increasingly complex visual patterns over increasingly larger spatial footprints. 
Our goal is to impose the desirable inductive bias for ViTs blocks of the hierarchical architecture. Thus, our explorations are directed to two paths with pure CNNs structure: \textit{Macro Design} and \textit{Micro Design}.

\subsection{Macro Design}
\label{sec:3_1}

%\begin{figure}[ht]
%\centering
%\includegraphics[width=\linewidth]{figs/Integrated schemes.pdf}
%\caption{Comparison for model generalization under different Integrated schemes. For fair
%comparison, all models have similar parameter and Flops.}
%\label{Fig:Integrated schemes}
%\end{figure}

% CNNs are good at capturing local structure features by using local receptive fields, shared weights, and spatial subsampling achieves some degree of shift, scale, and distortion invariance. Transformer has the merits of dynamic attention, global context, and better generalization. Therefore, the advantages of these two structures can form a good complement to a certain extent. How to merge different types of computing blocks to develop a competitive network architecture is worth exploring. In this section, we focus on how to combine the convolutions and Transformers effectively. Roughly speaking, how to stack different types of computational blocks together to form a better network.
\paragraph{Convolutional Stem.} In ViTs-based network design, the \textit{Stem} is concerned to extract the right inductive bias to following global attention modules. 
%After EarlyConv \cite{xiao2021early}, most ViT-based networks adopt the pure convolutional Stem in the early stage to effectively extract the local patterns. 
It is necessary that maintain enough effective receptive field (ERF) of CE to extract rich visual features at the early stage \cite{xiao2021early}. In consideration of parameter numbers and FLOPs budget, as well as we notice that $S0$ can be merged with the CE of $S1$, our final \textit{Stem}  consists of 4 Fused-MBConvs \cite{gupta2019efficientnet} layers and 1 regular convolutional layer. 
Specifically, $S0$ contains 2 Fused-MBConv with stride 2 and stride 1, respectively. The CE of $S1$ is consists of the same as $S0$ and followed by one 1×1 convolution at the end to match the channels and normalized by a layer normalization \cite{ba2016layer}. 
Another reason that we choose Fused-MBConv to compose \textit{Stem} is that EfficientNetV2 \cite{tan2021efficientnetv2} shows that Fused-MBConv is surprisingly effective in achieving better generalization and capacity as well as makes training convergence rate faster \cite{xiao2021early}. 
Besides, like EfficientNetV2,  the hidden state's expand ratios of Fused-MBConv and MBConv \cite{sandler2018mobilenetv2} are arranged to 1 or 2. Such a setting allows the convolutional \textit{Stem} to have fewer parameters and FLOPs without the loss of accuracy. Please refer to the supplementary materials for more details about Fused-MBConv and MBConv.

\paragraph{Convolutional Embedding.} 
In the following stages, $S2$, $S3$, and $S4$, each stage contains a CE block and a set of ViTs blocks. The CE block captures diverse deep representation with a convolutional inductive bias for subsequent attention modules. It is worth noting that EarlyConv \cite{xiao2021early} and CoAtNet \cite{dai2021coatnet} point out stacking convolutional structure may enhance the ViTs in early stage. However, as we show in Table \ref{tab:different_integrated_schemes}, we argue that CNNs is also able to represent the same, if not better, deep feature as ViTs in the deep layer of a network. Meanwhile,  maintaining the CNNs design of the embedding layer naturally introduces proper inductive bias to following ViTs and retains the sample-efficient learning property. For the same computation resource constraint consideration, the CE adopts only effective and efficient convolutions. The CE of $S2$, $S3$, and $S4$ adopts MBConv as the basic unit.

\subsection{Micro Design}
\label{sec:3_2}
\paragraph{Locally Enhanced Window Self-Attention.}
Previous work \cite{liu2021swin} restricts the attention calculation with a local window to reduce computational complexity from quadratic to linear. Meanwhile, to some extent, the local window injects some inductive bias into ViTs. Concurrently, a set of works \cite{wu2021cvt,wang2021pyramid} attempt to directly integrate CNNs into ViTs to bring in convolutional inductive bias, yet their computational complexity is still quadratic. 
We present a simple Locally Enhanced Window Self-Attention (LEWin) to take advantage of fast attention calculating and locality inductive bias. Inspired by CvT, we performed a \textit{convolutional projection} on input tokens in a local shift window. The \textit{convolutional projection} is implemented by a depth-wise separable convolution with kernel size 3 × 3, stride 1, and padding 1. The LEWin can be formulated as:
% \begin{equation}\label{...}
\begin{align}
&\bm{{x}^{i-1}} =  {\rm Flatten}\left ( {\rm Conv2D}\left ( {\rm Reshape}\left ( \bm{z^{i-1}}  \right ) \right ) \right ), \\
&\bm{\hat{z}^{i}}={\rm WHMSA}\left ({\rm LN}\left ( \bm{{x}^{i-1}} \right  )\right) + \bm{{x}^{i-1}},
\\
&\bm{z^{i}} = {\rm MLP}\left ( {\rm LN}\left ( \bm{\hat{z}^{i}} \right ) \right )+\bm{\hat{z}^{i}},
\\
&\bm{{x}^{i}} =  {\rm Flatten}\left ( {\rm Conv2D}\left ( {\rm Reshape}\left ( \bm{{z}^{i}} \right ) \right ) \right ), \\
&\bm{\hat{z}^{i+1}}={\rm SWMSA}\left ( {\rm LN}\left ( \bm{{x}^{i}} \right ) \right ) + \bm{{x}^{i}},
\\
&\bm{z^{i+1}} = {\rm MLP}\left ( {\rm LN} \left ( \bm{\hat{z}^{i+1}} \right ) \right )+\bm{\hat{z}^{i+1}}
\end{align}
% \end{equation}

where $\bm{x^i}$ is the unperturbed token in local window prior to the \textit{convolutional projection}, $\bm{\hat{z}^{i}}$ and $\bm{z^{i}}$ denote the output features of the WMSA or S-WMSA module and the MLP module for the $i-th$ block, respectively. W-MSA and SW-MSA define window-based multi-head self-attention based on the regular and shifted windows from SWin \cite{liu2021swin}, respectively. 

\subsection{CETNet Variants}
We consider three different network configurations for CETNet to compare with other ViTs backbones under similar model size and computation complexity conditions. By changing the base channel dimension and the number of ViTs blocks of each stage, we build three variants, tiny, small, and base models, namely CETNet-T, CETNet-S, and CETNet-B. For more detailed configurations, please refer to our supplementary materials.

\section{Experiments}
\label{sec:experiment}
To verify the ability of our CETNet as a general vision backbone.  We conduct experiments on ImageNet-1K  classification\cite{deng2009imagenet}, COCO object detection\cite{lin2014microsoft}, and ADE20K semantic segmentation\cite{zhou2019semantic}. In addition, comprehensive ablation studies are performed to validate the design of the proposed architecture.

\begin{table}[ht]
\centering
\caption{Comparison of image classification on ImageNet-1K for different models. The models are grouped based on the similar model size and computation complexity}
\resizebox{1.0\linewidth}{!}
{
\begin{tabular}{cccccccccccc}
\toprule
\multicolumn{12}{c}{\textbf{ImageNet-1K $ 224^2 $ trained model}} \\ 
\midrule
\multicolumn{1}{c|}{Model} &
  Params &
  FLOPs &
  \multicolumn{1}{c|}{Top-1(\%)} &
  Model &
  Params &
  FLOPs &
  \multicolumn{1}{c|}{Top-1(\%)} &
  Model &
  Params &
  FLOPs &
  Top-1(\%)  \\ 
\midrule
\multicolumn{1}{c|}{ResNet-50\cite{he2016deep}} & 25M & 4.1G & \multicolumn{1}{c|}{76.2} & ResNet-101\cite{he2016deep} & 45M & 7.9G   & \multicolumn{1}{c|}{77.4} & ResNet-152\cite{he2016deep} & 60M & 11.0G & 78.3 \\

\multicolumn{1}{c|}{RegNetY-4G\cite{radosavovic2020designing}} & 21M & 4.0G & \multicolumn{1}{c|}{80.0} & RegNetY-8G\cite{radosavovic2020designing} & 39M & 8.0G   & \multicolumn{1}{c|}{81.7} & RegNetY-16G\cite{radosavovic2020designing} & 84M & 16.0G & 82.9 \\

\multicolumn{1}{c|}{DeiT-S\cite{touvron2021training}}    & 22M & 4.6G & \multicolumn{1}{c|}{79.8} & PVT-M\cite{wang2021pyramid}      & 44M & 6.7G  & \multicolumn{1}{c|}{81.2} & DeiT-B\cite{touvron2021training}      & 87M & 17.5G  & 81.8 \\

\multicolumn{1}{c|}{PVT-S\cite{wang2021pyramid}}     & 25M & 3.8G & \multicolumn{1}{c|}{79.8} & T2T-19\cite{yuan2021tokens}     & 39M & 8.9G  & \multicolumn{1}{c|}{81.5} & PiT-B\cite{heo2021rethinking}        & 74M & 12.5G  & 82.0 \\

\multicolumn{1}{c|}{T2T-14\cite{yuan2021tokens}}    & 22M & 5.2G & \multicolumn{1}{c|}{81.5} & T2T$_t$ -19\cite{yuan2021tokens}     & 39M & 9.8G  & \multicolumn{1}{c|}{82.2} & T2T-24\cite{yuan2021tokens}      & 64M & 14.1G  & 82.3 \\

\multicolumn{1}{c|}{ViL-S\cite{zhang2021multi}}     & 25M & 4.9G & \multicolumn{1}{c|}{82.0} & ViL-M\cite{zhang2021multi}      & 40M & 8.7G  & \multicolumn{1}{c|}{83.3} & T2T$_t$ -24\cite{yuan2021tokens}    & 64M & 15.0G  & 82.6 \\

\multicolumn{1}{c|}{TNT-S\cite{han2021transformer}}     & 24M & 5.2G & \multicolumn{1}{c|}{81.3} & MViT-B\cite{fan2021multiscale}     & 37M & 7.8G  & \multicolumn{1}{c|}{81.0} & CPVT-B\cite{chu2021conditional}     & 88M & 17.6G  & 82.3 \\

\multicolumn{1}{c|}{CViT-15\cite{chen2021crossvit}}   & 27M & 5.6G & \multicolumn{1}{c|}{81.0} & CViT-18\cite{chen2021crossvit}    & 43M & 9.0G  & \multicolumn{1}{c|}{82.5} & TNT-B\cite{han2021transformer}      & 66M & 14.1G  & 82.8 \\

\multicolumn{1}{c|}{LViT-S\cite{li2021localvit}}    & 22M & 4.6G & \multicolumn{1}{c|}{80.8} & CViT$_c$-18\cite{chen2021crossvit}   & 44M & 9.5G  & \multicolumn{1}{c|}{82.8} & ViL-B\cite{zhang2021multi}      & 56M & 13.4G  &83.2 \\

\multicolumn{1}{c|}{CPVT-S\cite{chu2021conditional}}    & 23M & 4.6G & \multicolumn{1}{c|}{81.9} & Twins-B\cite{chu2021twins}    & 56M & 8.3G & \multicolumn{1}{c|}{83.2} & Twins-L\cite{chu2021twins}    & 99M & 14.8G  & 83.7 \\

\multicolumn{1}{c|}{Swin-T\cite{liu2021swin}}    & 29M & 4.5G & \multicolumn{1}{c|}{81.3} & Swin-S\cite{liu2021swin}     & 50M & 8.7G  & \multicolumn{1}{c|}{83.0} & Swin-B\cite{liu2021swin}     & 88M  & 15.4G  & 83.5 \\

\multicolumn{1}{c|}{CvT-13\cite{wu2021cvt}} & 20M & 4.5G & \multicolumn{1}{c|}{81.6} & CvT-21\cite{wu2021cvt} & 32M & 7.1G & \multicolumn{1}{c|}{82.5} & CETNet-B & 75M & 15.1G & \textbf{83.8} \\

\multicolumn{1}{c|}{CETNet-T} & 23M & 4.3G & \multicolumn{1}{c|}{\textbf{82.7}} & CETNet-S & 34M & 6.8G & \multicolumn{1}{c|}{\textbf{83.4}} &  &  &  &  \\

\bottomrule
\toprule

\multicolumn{12}{c}{\textbf{ImageNet-1K $ 384^2 $ finetuned model}} \\ 
\midrule
\multicolumn{1}{c|}{CvT-13\cite{wu2021cvt}} & 25M & 16.3G & \multicolumn{1}{c|}{83.0} & CvT-21\cite{wu2021cvt} & 32M & 24.9G   & \multicolumn{1}{c|}{83.3} & ViT-B/16\cite{dosovitskiy2020image} & 86M & 49.3G & 77.9 \\
\multicolumn{1}{c|}{T2T-14\cite{yuan2021tokens}} & 22M & 17.1G & \multicolumn{1}{c|}{83.3} & CViT$_c$-18\cite{chen2021crossvit}  & 45M & 32.4G   & \multicolumn{1}{c|}{83.9} & DeiT-B\cite{touvron2021training} & 86M & 55.4.3G & 83.1 \\
\multicolumn{1}{c|}{CViT$_c$-15\cite{chen2021crossvit}}   & 28M & 21.4G & \multicolumn{1}{c|}{83.5} & CETNet-S & 34M & 19.9G & \multicolumn{1}{c|}{\textbf{84.6}} & Swin-B\cite{liu2021swin}      & 88M & 47.0G  & 84.5 \\
\multicolumn{1}{c|}{CETNet-T}   & 24M & 12.5G & \multicolumn{1}{c|}{\textbf{84.2}} &  &  &  & \multicolumn{1}{c|}{} & CETNet-B  & 75M & 44.50G  & \textbf{84.9} \\
\bottomrule

\end{tabular}
}
\label{tab:classification}
\end{table}

\subsection{Image Classification}
\label{sec:experiment_setting}
\subsubsection{Settings} For Image classification, we compare different methods on ImageNet-1K\cite{deng2009imagenet}, with about 1.3M images and 1K classes, the training set, and the validation set containing 1.28M images and 50K images respectively. The top 1 accuracy on the validation set is reported to show the capacity of our CETNet.
For a fair comparison, the experiment setting follows the training strategy in \cite{liu2021swin}. All model variants are trained for 300 epochs with a batch size of 1024 or 2048 and using a cosine decay learning rate scheduler with 20 epochs of linear warm-up. We adopt an AdamW\cite{kingma2014adam} optimizer with an initial learning rate of 0.001, and a weight decay of 0.05 is used. We use the same data augmentation methods, and regularization strategies used in \cite{liu2021swin} for training. All models are trained with $224\times224$ input size, while a center crop is used during evaluation on the validation set. When fine-tuning on $384\times384$ input, we train the models for 30 epochs with a learning rate of 2e-5, batch size of 512, and no center crop for evaluation.

\subsubsection{Results} Table \ref{tab:classification} presents comparisons to other models on the image classification task, and the models are split into three groups based on the similar model size (Params) and computation complexity (FLOPs). It can be seen from the table, our models consistently exceed other methods by large margins. More specifically, CETNet-T achieves 82.7\% Top-1 accuracy with only 23M parameters and 4.3G FLOPs, surpassing CvT-13, Swin-T, and DeiT-S by 1.1\%, 1.4\% and 2.9\% separately. For small and base models, our CETNet-S and  CETNet-B still achieve better performance with comparable Params and FLOPs. Compared with the state-of-the-art CNNs RegNet\cite{radosavovic2020designing} which also trained with input size $224\times224$ with out extra data, our CETNet achieves better accuracy. On the 384 × 384 input, Our CETNet's performance is still better than other Backbone, achieve 84.2\%, 84.6\% and 84.9\% respectively, well demonstrates the powerful learning capacity of our CETNet. As a result, we can conclude that an effective combination of CNNs and ViTs can optimize well in small(or middle) scale datasets.

\begin{table}[ht]
\centering
\caption{Object detection and instance segmentation performance on COCO 2017 validation set with the Mask R-CNN framework. The FLOPs are measured at a resolution of 800 × 1280, and the backbones are pre-trained on the ImageNet-1K}
\resizebox{1.0\linewidth}{!}{

\begin{tabular}{ccccccccc|cccccc}
\toprule
\multicolumn{9}{c|}{Mask R-CNN 1× schedule} & \multicolumn{6}{c}{Mask R-CNN 3× schedule} \\ 
%\hline
\midrule
Backbone & Params & FLOPs & $AP^b$ & $AP^{b}_{50}$ & $AP^{b}_{75}$ & $AP^{m}$ & $AP^{m}_{50}$ & $AP^{m}_{75}$ & $AP^{b}$ & $AP^{b}_{50}$ & $AP^{b}_{75}$ & $AP^{m}$ & $AP^{m}_{50}$ & $AP^{m}_{70}$ \\ 
%\hline
\midrule

Res50\cite{he2016deep} & 44M & 260G & 38 & 58.6 & 41.4 & 34.4 & 55.1 & 36.7 & 41 & 61.7 & 44.9 & 37.1 & 58.4 & 40.1 \\
PVT-S\cite{wang2021pyramid} & 44M & 245G & 40.4 & 62.9 & 43.8 & 37.8 & 60.1 & 40.3 & 43 & 65.3 & 46.9 & 39.9 & 62.5 & 42.8 \\
ViL-S\cite{zhang2021multi} & 45M & 218G & 44.9 & 67.1 & 49.3 & \textbf{41.0} & 64.2 & \textbf{44.1} & \textbf{47.1} & 68.7 & 51.5 & \textbf{42.7} & \textbf{65.9} & \textbf{46.2} \\
TwinsP-S\cite{chu2021twins} & 44M & 245G & 42.9 & 65.8 & 47.1 & 40.0 & 62.7 & 42.9 & 46.8 & \textbf{69.3} & \textbf{51.8} & 42.6 & 66.3 & 46 \\
Twins-S\cite{chu2021twins} & 44M & 228G & 43.4 & 66.0 & 47.3 & 40.3 & 63.2 & 43.4 & 46.8 & 69.2 & 51.2 & 42.6 & 66.3 & 45.8 \\
Swin-T\cite{liu2021swin} & 48M & 264G & 43.7 & 64.6 & 46.2 & 39.1 & 61.6 & 42.0 & 46 & 68.2 & 50.2 & 41.6 & 65.1 & 44.8 \\
CETNet-T & 43M & 261G & \textbf{45.5} & \textbf{67.7} & \textbf{50.0} & 40.7 & \textbf{64.4} & 43.7 & 46.9 & 67.9 & 51.5 & 41.6 & 65 & 44.7 \\ 
%\hline
\midrule

Res101\cite{he2016deep} & 63M & 336G & 40.4 & 61.1 & 44.2 & 36.4 & 57.7 & 38.8 & 42.8 & 63.2 & 47.1 & 38.5 & 60.1 & 41.3 \\
X101-32\cite{xie2017aggregated} & 63M & 340G & 41.9 & 62.5 & 45.9 & 37.5 & 59.4 & 40.2 & 42.8 & 63.2 & 47.1 & 38.5 & 60.1 & 41.3 \\
PVT-M\cite{wang2021pyramid} & 64M & 302G & 42.0 & 64.4 & 45.6 & 39.0 & 61.6 & 42.1 & 42.8 & 63.2 & 47.1 & 38.5 & 60.1 & 41.3 \\
ViL-M\cite{zhang2021multi} & 60M & 261G & 43.4 & -- & -- & 39.7 & -- & -- & 44.6 & 66.3 & 48.5 & 40.7 & 63.8 & 43.7 \\
TwinsP-B\cite{chu2021twins} & 64M & 302G & 44.6 & 66.7 & 48.9 & 40.9 & 63.8 & 44.2 & 47.9 & 70.1 & 52.5 & 43.2 & 67.2 & 46.3 \\
Twins-B\cite{chu2021twins} & 76M & 340G & 45.2 & 67.6 & 49.3 & 41.5 & 64.5 & 44.8 & 48 & 69.5 & 52.7 & 43 & 66.8 & 46.6 \\
Swin-S\cite{liu2021swin} & 69M & 354G & 44.8 & 66.6 & 48.9 & 40.9 & 63.4 & 44.2 & 48.5 & \textbf{70.2} & 53.5 & \textbf{43.3} & \textbf{67.3} & \textbf{46.6} \\
CETNet-S & 53M & 315G & \textbf{46.6} & \textbf{68.7} & \textbf{51.4} & \textbf{41.6} & \textbf{65.4} & \textbf{44.8} & \textbf{48.6} & 69.8 & \textbf{53.5} & 43 & 66.9 & 46 \\ 
%\hline
\midrule

X101-64\cite{xie2017aggregated} & 101M & 493G & 42.8 & 63.8 & 47.3 & 38.4 & 60.6 & 41.3 & 44.4 & 64.9 & 48.8 & 39.7 & 61.9 & 42.6 \\
PVT-L\cite{wang2021pyramid} & 81M & 364G & 42.9 & 65.0 & 46.6 & 39.5 & 61.9 & 42.5 & 44.5 & 66.0 & 48.3 & 40.7 & 63.4 & 43.7 \\
ViL-B\cite{zhang2021multi} & 76M & 365G & 45.1 & -- & -- & 41.0 & -- & -- & 45.7 & 67.2 & 49.9 & 41.3 & 64.4 & 44.5 \\
TwinsP-L\cite{chu2021twins} & 81M & 364G & 45.4 & -- & -- & 41.6 & -- & -- & -- & -- & -- & -- & -- & -- \\
Twins-L\cite{chu2021twins} & 111M & 474G & 45.9 & -- & -- & 41.6 & -- & -- & -- & -- & -- & -- & -- & -- \\
Swin-B\cite{liu2021swin} & 107M & 496G & 46.9 & -- & -- & 42.3 & -- & -- & 48.5 & \textbf{69.8} & 53.2 & \textbf{43.4} & 66.8 & \textbf{46.9} \\
CETNet-B & 94M & 495G & \textbf{47.9} & \textbf{70.3} & \textbf{53.0} & \textbf{42.5} & \textbf{67.2} & \textbf{45.6} & \textbf{48.6} & 69.5 & \textbf{53.7} & 43.1 & \textbf{66.9} & 46.4 \\ 
\bottomrule

\end{tabular}
}

\label{tab:detection}
\end{table}

% \vspace*{-0.6cm}
\subsection{Object Detection and Instance Segmentation}
\subsubsection{Settings} For object detection and instance segmentation experiments, we evaluate our CETNet with the Mask R-CNN\cite{he2017mask} framework on COCO 2017, which contains over 200K images with 80 classes. The models pre-trained on the ImageNet-1K dataset are used as visual backbones. We follow the standard to use two training schedules, 1× schedule(12 epochs with the learning rate decayed by 10× at epochs 8 and 11) and 3× schedule(36 epochs with the learning rate decayed by 10× at epochs 27 and 33). We utilize the multi-scale training strategy\cite{carion2020end}, resizing the input that the shorter side is between 480 and 800 while the longer side is no more than 1333. AdamW\cite{kingma2014adam} optimizer with initial learning rate of 0.0001, weight decay of 0.05. All models are trained on the 118K training images with a batch size of 16, and the results are reported on 5K images in COCO 2017 validation set.

% save original \intextsep
\newlength{\oldintextsep}
\setlength{\oldintextsep}{\intextsep}

\setlength\intextsep{-5pt}
% \begin{table}[ht]
\begin{wraptable}{r}{0.5\textwidth}
\centering
% \vspace{-32pt}
\caption{Comparison of semantic segmentation on ADE20K with the Upernet framework, both single and multi-scale evaluations are reported in the last two columns. FLOPs are calculated with a resolution of 512×2048, and the backbones are pre-trained on the ImageNet-1K}
% \vspace{5pt}
\resizebox{1\linewidth}{!}{
\begin{tabular}{ccccc}
\toprule
\multicolumn{5}{c}{Upernet 160k trained models}                       \\ \hline
\multicolumn{1}{l}{Backbone} & \multicolumn{1}{l}{Params} & \multicolumn{1}{l}{FLOPs} & \multicolumn{1}{l}{mIoU(\%)} & \multicolumn{1}{l}{MS mIoU(\%)} \\ \hline
TwinsP-S\cite{chu2021twins} & 54.6M & 919G  & 46.2     & 47.5           \\
Twins-S\cite{chu2021twins}  & 54.4M & 901G & 46.2     & 47.1            \\
Swin-T\cite{liu2021swin}   & 59.9M & 945G & 44.5     & 45.8             \\
CETNet-T   & 53.2M & 935G & \textbf{46.5} & \textbf{47.9} \\ \hline
Res101\cite{he2016deep}   & 86.0M  & 1029G & --            & 44.9       \\
TwinsP-B\cite{chu2021twins} & 74.3M & 977G & 47.1          & 48.4       \\
Twins-B\cite{chu2021twins}  & 88.5M & 1020G & 47.7          & 48.9      \\
Swin-S\cite{liu2021swin}   & 81.3M & 1038G & 47.6          & 49.5       \\
CETNet-S   & 63.4M & 990G & \textbf{48.9} & \textbf{50.6} \\ \hline
TwinsP-L\cite{chu2021twins} & 91.5M & 1041G & 48.6          & 49.8      \\
Twins-L\cite{chu2021twins}  & 133.0M & 1164G & 48.8          & 50.2     \\
Swin-B\cite{liu2021swin}   & 121.0M & 1188G & 48.1          & 49.7      \\
CETNet-B   & 106.3M & 1176G & \textbf{50.2} & \textbf{51.6} \\ 
\bottomrule
\end{tabular}
}
\label{tab:semantic }

\end{wraptable}
% \end{table}
\setlength\intextsep{-10pt}
\setlength{\oldintextsep}{\intextsep}

\subsubsection{Results} As shown in Table \ref{tab:detection}, the results of the Mask R-CNN framework show that our CETNet variants clearly outperform all counterparts with 1$\times$ schedule. In detail, our CETNet-T outperforms Swin-T by +1.8 box AP, +1.6 mask AP. On the small and base configurations, the performance gain can also be achieved with +1.8 box AP, +0.7 mask AP and +1.0 box AP, +0.2 mask AP respectively. For 3× schedule, our model, CETNet-S and CETNet-B, can achieve competitive results with lower Params and FLOPs than the current state-of-the-art ViTs methods in small and base scenarios. 
However, like SWin, CETNet-T can't beat the SOTA performance. Also, in the small and base configurations, we notice the variants of our CETNet and SWin do not improve the performance. We conjecture such inferior performance may be because of insufficient data.

\setlength{\oldintextsep}{\intextsep}

\subsection{Semantic Segmentation}
\subsubsection{Settings} We further use the pre-trained models as the backbone to investigate the capability of our models for Semantic Segmentation on the ADE20K\cite{zhou2019semantic} dataset. ADE20K is a widely-used semantic segmentation dataset that contains 150 fine-grained semantic categories, with 20,210, 2,000, and 3,352 images for training, validation, and testing, respectively. We follow previous works\cite{liu2021swin} to employ UperNet\cite{xiao2018unified} as the basic framework and follow the same setting for a fair comparison. In the training stage, we employ the AdamW\cite{kingma2014adam} optimizer and set the initial learning rate to 6e-5 and use a polynomial learning rate decay, and the weight decay is set to 0.01 and train Upernet 160k iterations with batch size of 16. The data augmentations adopt the default setting in mmsegmentation of random horizontal flipping, random re-scaling within ratio range [0.5, 2.0], and random photometric distortion. Both single and multi-scale inference are used for evaluation. 
\subsubsection{Results} In Table \ref{tab:semantic }, we list the results of Upernet 160k trained model on ADE20K. It can be seen that our CETNet models significantly outperform previous state-of-the-arts under different configurations. In details, CETNet-T achieves 46.5 mIoU and 47.9 multi-scale tested mIoU, +2.0 and + 2.1 higher than Swin-T with  similar computation cost. On the small and base configurations, CETNet-S and CETNet-B still achieve  +1.3 and +2.1 higher mIOU and +1.1 and +1.9 multi-scale tested mIoU than the Swin counterparts. The performance gain is promising and demonstrates the effectiveness of our CETNet design.

\subsection{Ablation Study}
In this section, we ablate the critical elements of the proposed CETNet backbone using ImageNet-1K image classification. The experimental settings are the same as the settings in Sec. \ref{sec:experiment_setting}.
For the attention mechanism, when replacing the micro design LEWin self-attention with the origin shifted window self-attention, the performance dropped 0.2\%, demonstrating the effectiveness of our micro design. Furthermore, we find the “deep-narrow” architecture is better than the “shallow-wide” counterpart. Specifically, the deep-narrow model with [2,2,18,2] Transformer blocks for four stages with the base channel dimensions $D=64$ and the shallow-wide model with [2,2,6,2] blocks for four stages with $D=96$. As we can see from the 1st and 3rd rows, even with larger Params and FLOPs, the shallow-wide model performs worse than the deep-narrow design. 
The CE module is the key element in our models. To verify the effectiveness CE module, we compare it with the existing method used in hierarchical ViTs backbones, including the patch embedding and patch merging modules in SWin \cite{liu2021swin} and convolutional token embedding modules described in CvT \cite{wu2021cvt}. For a fair comparison, we use the shallow-wide design mentioned above and apply these three methods in all these models with all other factors kept the same. As shown in the last three rows of Table \ref{tab:ablation}, our CE module performs better than other existing methods.

\setlength\intextsep{15pt}
\begin{table}[]
\centering
\caption{Ablation study of CETNet's key elements. `Without LEAtten' denotes replacing the micro design LEWin self-attention with the shifted window self-attention in SWin \cite{liu2021swin}. The `D' and `S' represents the deep-narrow and the shallow-wide model, respectively. The 1st and 3rd rows are the baseline `D' and `S' models.
%The `Deep-narrow` and `Shallow-wide` model with [2,2,18,2] and [2,2,6,2] Transformer blocks for four stages and the base channel dimensions D set to 64 and 96, respectively.
%Shallow-Wide architecture is a variant of CETNet-T that has [2,2,6,2] Transformer blocks for four stages and the base channel dimensions D as 96
}
\begin{tabular}{cccc}
\toprule
Models                                                     & Param & FLOPs  & Top-1(\%)      
\\ 
\hline
CETNet-T (D with CE)                                                    & 23.4M   & 4.3G & \textbf{82.7} 
\\  
Without LEAtten (D with CE)                                              & 23.3M   & 4.2G    & 82.5          
\\
Shallow-wide (S with CE)                                   & 29M     & 4.6G  & 82.5          
\\ 
Patch embedding+patch merging \cite{liu2021swin} (S)                              & 27M     & 4.6G  & 81.5          
\\ 
Convolutional token embedding \cite{wu2021cvt} (S)                               & 27M     & 4.6G  & 82.0          
\\ 
\bottomrule
\end{tabular}
\label{tab:ablation}
\end{table}
\setlength{\oldintextsep}{\intextsep}

\section{Investigating the role of Convolutional Embedding}

This section systematically understands how the hybrid CNNs/ViTs network benefits from the CE. As we need carefully make the trade-off among the computation budget, the model generalization, and the model capacity, the CE, two design choices are mainly explored in this work: the number of CNNs layers of CE and the basic unit of CE. Besides, we show the superiority of CE via integrate CE into 4 popular hierarchical ViTs models. Finally, we further investigate the role of CNNs in hybrid CNNs/ViTs network design.
Except for specific declaration, all configuration is follow the Image Classification setting as section \ref{sec:experiment}.

\subsection{Effect of the Stacking Number}
% 1- -7/ -5- -10
\setlength\intextsep{-4pt}
\begin{wraptable}{r}{0.5\textwidth}
% \vspace{-30pt}
\centering
\caption{Performance of model with different layer numbers of CE. The baseline is slightly modified from SWin-Tiny \cite{liu2021swin}}
%  \vspace{5pt}
\resizebox{1\linewidth}{!}{
\begin{tabular}{cccc}
\toprule
Models                                                     & Param & FLOPs  & Top-1(\%)      \\ 
\midrule
Swin-T\cite{liu2021swin}                                                 & 29M     & 4.5G    & 81.3           \\
1-layer CE                                                 & 28.1M     & 4.5G    & 81.8           \\
3-layer CE                                                 & 28.1M     & 4.6G    & 82.2           \\
5-layer CE                                                 & 29.1M   & 4.9G  & 82.4          \\
7-layer CE                                                 & 30.1M   & 5.4G  & 82.6          \\ 
\bottomrule
\end{tabular}
\label{tab:Table_numberOfCE}
}
\end{wraptable}
\setlength\intextsep{-10pt}
\setlength{\oldintextsep}{\intextsep}

We first explore how large effective receptive field (ERF) affect the computation budget and the performance of the network.
As we mentioned in section \ref{sec:3_1}, the $S0$ and the CE of $S1$ are combined as one whole CNNs stem. We refer it as the first CE ($\text{CE}^{1st}$) in rest of this section, $\text{CE}^{2nd}$, $\text{CE}^{3rd}$, and $\text{CE}^{4th}$ represent the CE of $S2$, $S3$, $S4$, respectively.
To explore the effect of the stacking number of the basic unit, we slightly modify the SWin-Tiny \cite{liu2021swin} model by replacing its patch embedding and patch merging modules with pure CNNs. We choose MBConv as the basic unit of CE and gradually increase the layer number of CE from 1 to 3, 5, and 7. As shown in Table \ref{tab:Table_numberOfCE}, as the CE contains more MBConvs, the Param and FLOPs grow, and the performance increases from 81.8\% to 82.6\%. The 3-layer CE model has slightly higher FLOPs than the 1-layer CE (4.48G vs. 4.55G) with near the same Param (28.13M vs. 28.11M). Besides, it is worth noticing that the performance grows negligibly from 5-layer CE to 7-layer CE. Thus, we employ 5-layer setting,  offering large ERF \cite{yu2015multi,tan2021efficientnetv2}, as CETNet's final configuration.

\subsection{Effect of Different CNNs Blocks}
\setlength\intextsep{-5pt}
%\begin{table}[ht]
\begin{wraptable}{r}{0.5\textwidth}
% \vspace{-31pt}
\centering
\caption{Transfer some popular CNNs blocks to CETNet-T, including the DenseNet block, ShuffleNet block, ResNet block, SeresNet block, and GhostNet block. PureMBConv represents to use MBConv block to replace the Fused-MBConv block in the early stage of CETNet-T}
%  \vspace{5pt}
\resizebox{1\linewidth}{!}{
\begin{tabular}{cccc}
\toprule
CNNs type & Params & FLOPs & Top-1(\%) \\ 
\hline
CETNet-T & 23.4M & 4.3G & \textbf{82.7} \\
PureMBConvs\cite{sandler2018mobilenetv2} & 23.4M & 4.2G & 82.6 \\
GhostNetConvs\cite{han2020ghostnet} & 24.6M & 4.3G & 82.6 \\ 
DenseNetConvs\cite{huang2017densely} & 22.3M & 4.4G & 82.5 \\
ShuffleNetConvs\cite{zhang2018shufflenet} & 23.4M & 4.3G & 82.4 \\
ResNetConvs\cite{he2016deep} & 23.8M & 4.3G & 82.0 \\
SeresNetConvs\cite{hu2018squeeze} & 24.0M & 4.3G & 82.0 \\
\bottomrule
\end{tabular}
\label{tab:table_different_convs}
%\end{table}
}
\end{wraptable}
\setlength\intextsep{-10pt}
\setlength{\oldintextsep}{\intextsep}

We next explore how well convolutional inductive bias inject by different CNNs architectures.
The CE layers aim to offer rich features for later attention modules. CNNs' effective receptive field (ERF) determines the information it can cover and process. As Luo \textit{et al.} mentioned, stacking more layers, subsampling, and CNNs with large ERF, such as dilated convolution \cite{yu2015multi} can enlarge the ERF of CNNs networks. For find an efficient basic unit for CE, we use the CETNet-Tiny (CETNet-T) model as a baseline and replace the CE with some recent CNNs building blocks, such as MBConv \cite{sandler2018mobilenetv2}, DenseNet \cite{huang2017densely}, ShuffleNet \cite{zhang2018shufflenet}, ResNetConvs\cite{he2016deep}, SeresNetConvs \cite{hu2018squeeze}, and GhostNetConvs \cite{han2020ghostnet}. All candidate convolutions layers stack to 5 layers based on previous finding. 
For a fair comparison, all model variants are constructed to have similar parameter numbers and FLOPs. 
As we can see from Table \ref{tab:table_different_convs}, when replaced the Fused-MBConvs in the early stage of CETNet-T by MBConvs, the top-1 accuracy increased 0.1\%, and we observed a 12 percent decrease in training speed. Also, CETNet-T and PureMB model achieve higher performance than other candidate convolutions. We argue that may be the internal relation between depthwise convolution and ViTs as pointed out by CoAtNet \cite{dai2021coatnet}, which is further verified by the GhostNet and shuffleNet model, which archive 82.6\% and 82.4\% top-1 accuracy. Besides, we noticed that under that CNNs help ViTs see better perspective, in CE case, dense connections in convolutions may not necessarily hurt performance. Since the result shows that our DenseNet model can also archive 82.5\% top-1 accuracy, which is comparable with the performance of CETNet-T, PureMB, GhostNet, and shuffleNet model. However, ResNet and SeresNet show inferior performance. We conjecture that the basic units have different ERF with the same stacking number.

\setlength\intextsep{-5pt}
\begin{wraptable}{R}{0.5\textwidth}
% \vspace{-31pt}
%\begin{table}[ht]
\centering
\caption{Generalize the CE module to 4 ViT backbones. All models are trained on ImageNet-1K dataset and compared with the original model under the same training scheme. Depths indicate the number of Transformer layers of each stage. FLOPs are calculated with a resolution of $224 \times 224$}
% \vspace{5pt}
\resizebox{1\linewidth}{!}{
\begin{tabular}{c|cccccc}
\toprule
Framework                &Models   & Channels & Depths                                & Param & FLOPs  & Top-1         \\ 
\midrule
\multirow{2}{*}{CvT\cite{wu2021cvt}}  & CvT-13   & 64       & {[}1, 2, 10{]}                        & 20.0M     & 4.5G  & 81.6          \\
                      & CE-CvT-13 & 64       & {[}1, 2, 10{]}                        & 20.0M     & 4.4G  & \textbf{82.1}(\textcolor{red}{0.5$\uparrow$}) \\ 
					  \midrule
\multirow{2}{*}{PVT\cite{wang2021pyramid}}  & PVT-S    & 64       & {[}3, 4, 6, 3{]}                      & 24.5M   & 3.8G  & 79.8          \\
                      & CE-PVT-S & 64       & {[}3, 4, 4, 3{]}                      & 22.6M   & 3.7G  & \textbf{81.1}(\textcolor{red}{1.3$\uparrow$}) \\ 
					  %\cline{2-7} 
                      \midrule
\multirow{6}{*}{SWin\cite{liu2021swin}} & Swin-T   & 96       & {[}2, 2, 6, 2{]}                      & 28.3M   & 4.5G  & 81.3          \\
                      & CE-Swin-T & 96       & {[}2, 2, 4, 2{]}                      & 23.4M   & 4.3G  & \textbf{82.5}(\textcolor{red}{1.2$\uparrow$}) \\ 
					  %\cline{2-7} 
                      & Swin-S   & 96       & {[}2, 2, 18, 2{]}                     & 50.0M   & 8.7G  & 83.0          \\
                      & CE-Swin-S & 96       & {[}2, 2, 16, 2{]}                     & 48.2M   & 8.8G  & \textbf{83.6}(\textcolor{red}{0.6$\uparrow$}) \\ 
					  %\cline{2-7} 
                      & Swin-B   & 128      & {[}2, 2, 18, 2{]}                     & 88.0M   & 15.4G  & 83.3          \\
                      & CE-Swin-B & 128      & {[}2, 2, 16, 2{]}                     & 85.2M   & 15.5G  & \textbf{84.0}(\textcolor{red}{0.7$\uparrow$}) \\ 
					  \midrule
\multirow{6}{*}{CSWin\cite{dong2021cswin}} & CSWin-T  & 64 & \multicolumn{1}{l}{{[}1, 2, 21, 1{]}} & 23.0M   & 4.3G  & 82.7          \\
                      & CE-CSWin-T & 64       & \multicolumn{1}{l}{{[}1, 2, 20, 1{]}} & 21.6M   & 4.2G  & \textbf{83.6}(\textcolor{red}{0.9$\uparrow$}) \\ 
					  %\cline{2-7} 
                      & CSWin-S  & 64       & \multicolumn{1}{l}{{[}2, 4, 32, 2{]}} & 35.0M     & 6.9G  & 83.6          \\
                      & CE-CSWin-S & 64       & \multicolumn{1}{l}{{[}2, 4, 31, 2{]}} & 33.9M   & 6.6G  & \textbf{84.1}(\textcolor{red}{0.5$\uparrow$}) \\ 
					  %\cline{2-7} 
                      & CSWin-B  & 96       & \multicolumn{1}{l}{{[}2, 4, 32, 2{]}} & 78.0M     & 15.0G & 84.2          \\
                      & CE-CSWin-B & 96 & \multicolumn{1}{l}{{[}2, 4, 31, 2{]}} & 75.8M & 14.7G & \textbf{84.7}(\textcolor{red}{0.5$\uparrow$}) \\ 
					  \bottomrule
\end{tabular}
}

\label{tab:table_CE_into_diff_ViTs}
% \end{table}
\end{wraptable}
\setlength\intextsep{-10pt}
\setlength{\oldintextsep}{\intextsep}

\subsection{Generalization of CE}
Then, we attempt to generalize the CE design to more ViTs backbones of CV. Here, we apply our CE design, 5-layer Fused-MBConv of $\text{CE}^{1st}$, and 5-layer MBConv of $\text{CE}^{2nd}$, $\text{CE}^{3rd}$, and $\text{CE}^{4th}$ respectively, to 4 prevalent backbones, CvT \cite{wu2021cvt}, PVT\cite{wang2021pyramid}, SWin \cite{liu2021swin}, and CSWin \cite{dong2021cswin}. For a fair comparison, we slightly change the structure, removing some ViTs blocks, of 4 models to keeps their parameter numbers and FLOPs maintaining the similar level as their original version. Also, we modify the small-scale model variant CvT-13 and PVT-S of CvT and PVT.
As shown in Table \ref{tab:table_CE_into_diff_ViTs}, those modified models outperform the original model 0.5\% and 1.3\% separately.
Furthermore, when introducing our design into SWin and CSWin, the top-1 accuracy of all counterparts is improved even under lower parameter numbers and FLOPs scenarios. For details, the modified models of Swin counterparts gain 1.2\%, 0.6\% and 0.7\%, and the CSwin counterparts gain 0.9\%, 0.5\% and 0.5\% respectively. Those results demonstrated that CE could be easily integrated with other ViT models and significantly improve the performance of those ViT models.

\subsection{Understanding the Role of CNNs in Hybrid CNNs/ViTs Design}
Finally, we explore how well CNNs in the deep layer of the hybrid CNNs/ViTs network improves ViTs. Previous works \cite{xiao2021early,dai2021coatnet,marquardt2004token} show the \textit{shallow} CNNs structure is enough to bring the convolutional inductive bias to all following ViTs blocks. However, one may notice that the $\text{CE}^{2nd}$, $\text{CE}^{3rd}$, and $\text{CE}^{4th}$ are \textbf{not} locate the \textit{shallow} layer of network. To fully understand: 1) whether CNNs in the deep layer enhances the inductive bias for subsequent ViTs blocks; 2) how hybrid CNNs/ViTs design affects the final performance of the network. We conduct the following experiments. From macro view, CETNet can be view as `C-T-C-T-C-T-C-T', where C and T denote CE and ViTs blocks respectively, where $\text{CE}^{1st}$ is Fused-MBConv, $\text{CE}^{2nd}$, $\text{CE}^{3rd}$, and $\text{CE}^{4th}$ are MBConv. We conduct three main experiments: \textbf{CNNs to ViTs}, \textbf{ViTs to CNNs}, and \textbf{Others}. In \textbf{CNNs to ViTs} group, we gradually replace the convolutions with transformers. In \textbf{ViTs to CNNs} group, we do the reverse. As we can see, only adopting CNNs in \textit{early} stage is not optimal. In addition, all hybrid models outperform the pure ViTs model in \textbf{CNNs to ViTs}. Besides, in comparison with \textbf{ViTs to CNNs}, one may notice that in deep layer architecture with more ViTs is superior to more CNNs. In addition, we have: $\text{hybrid CNNs/ViTs} \ge \text{pure ViT} \ge \text{pure CNNs}$, in deep layer of network. In \textbf{Others} group, we further list some variants' experiment results to the audience and hope that any possible insights may raise a rethinking of the hybrid CNNs/ViTs network design. 

\begin{comment}
\begin{wraptable}{r}{6.0cm}
\centering
\caption{Comparison of different Integrated schemes of CNNs and ViTs. "Arch" is a abbreviation of architecture. C represents MBConvs(Fused-MBConvs in early stage), and T represents the ViTs block mentioned in section \ref{sec:3_2}}
%  \vspace{5pt}
\resizebox{0.95\linewidth}{!}{
\begin{tabular}{cccc}
\toprule
Arch & Param & FLOPs & Top-1 \\ \hline
C-T-C-T-C-T-C-T & 23.4M & 4.3G & 82.7 \\
C-C-C-T-C-T-C-T & 23.5M & 4.3G & 82.7 \\
C-T-C-T-C-T-T-T & 24.0m & 4.2G & 82.8 \\
C-C-C-C-T-T-T-T & 25.5G & 4.4G & 81.8 \\
T-T-T-T-C-C-C-C & 23.4M & 4.8G & 76.3 \\
T-C-T-C-T-C-T-C & 24.5M & 4.2G & 79.8 \\ 
\bottomrule
\toprule
C-T-C-T-C-T-C-T & 23.4M & 4.3G & 82.7 \\
C-T-C-T-C-T-T-T & 24.0M & 4.2G & 82.8 \\
C-T-C-T-T-T-T-T & 24.1M & 4.2G & 82.5 \\
C-T-T-T-T-T-T-T & 24.1M & 4.4G & 82.3 \\
T-T-T-T-T-T-T-T & 24.3M & 4.2G & 80.1 \\ 
\bottomrule
\toprule
\multicolumn{1}{l}{C-T-C-T-C-T-C-T} & \multicolumn{1}{l}{23.4M} & \multicolumn{1}{l}{4.3G} & \multicolumn{1}{l}{82.7} \\
\multicolumn{1}{l}{C-T-C-T-C-T-C-C} & \multicolumn{1}{l}{23.7M} & \multicolumn{1}{l}{4.2G} & \multicolumn{1}{l}{82.0} \\
\multicolumn{1}{l}{C-T-C-T-C-C-C-C} & \multicolumn{1}{l}{24.4M} & \multicolumn{1}{l}{4.2G} & \multicolumn{1}{l}{79.6} \\
\multicolumn{1}{l}{C-T-C-C-C-C-C-C} & \multicolumn{1}{l}{24.3M} & \multicolumn{1}{l}{4.2G} & \multicolumn{1}{l}{79.2} \\
\multicolumn{1}{l}{C-C-C-C-C-C-C-C} & \multicolumn{1}{l}{24.6M} & \multicolumn{1}{l}{5.1G} & \multicolumn{1}{l}{79.0} \\ 
\bottomrule
\end{tabular}
\label{tab:hybrid}
}
\end{wraptable}
\end{comment}

\setlength\intextsep{10pt}
\begin{table}[h]
\centering
\caption{Comparison of different hybrid CNNs/ViTs designs. `Arch' represents architecture for short. C represents MBConvs(Fused-MBConvs in the early stage), and T represents the ViTs block mentioned in section \ref{sec:3_2}}
\resizebox{1.0\linewidth}{!}{
\begin{tabular}{cccc|ccccc|ccccc}
%\cline{1-4} \cline{6-9} \cline{11-14}
\toprule
 \multicolumn{4}{c|}{\textbf{CNNs to ViTs}} &  \multicolumn{5}{c|}{\textbf{ViTs to CNNs}} & \multicolumn{5}{c}{\textbf{Others}} \\ 
\midrule
Arch & Param & FLOPs & Top-1 &  & Arch & Param & FLOPs & Top-1 &  & Arch & Param & FLOPs & Top-1 \\ 
%\cline{1-4} \cline{6-9} \cline{11-14} 
\midrule
 C-T-C-T-C-T-C-T & 23.4M & 4.3G & 82.7 &  & C-T-C-T-C-T-C-T & 23.4M & 4.3G & 82.7 & & C-T-C-T-C-T-C-T & 23.4M & 4.3G & 82.7  \\
 C-T-C-T-C-T-T-T & 24.0M & 4.2G & 82.8 &  & C-T-C-T-C-T-C-C & 23.7M & 4.2G & 82.0 & & C-C-C-T-C-T-C-T & 23.5M & 4.3G & 82.7  \\
 C-T-C-T-T-T-T-T & 24.1M & 4.2G & 82.5 &  & C-T-C-T-C-C-C-C & 24.4M & 4.2G & 79.6 & & C-C-C-C-T-T-T-T & 25.5M & 4.4G & 81.8  \\
 C-T-T-T-T-T-T-T & 24.1M & 4.4G & 82.3 &  & C-T-C-C-C-C-C-C & 24.3M & 4.2G & 79.2 & & T-T-T-T-C-C-C-C & 23.4M & 4.8G & 76.3  \\
 T-T-T-T-T-T-T-T & 24.3M & 4.2G & 80.1 &  & C-C-C-C-C-C-C-C & 24.6M & 5.1G & 79.0 & & T-C-T-C-T-C-T-C & 24.5M & 4.2G & 79.8  \\
% T-C-T-C-T-C-T-C & 24.5M & 4.2G & 79.8 &  &  &  &  &  &  &  &  &  &  \\ %\cline{1-4} \cline{6-9} \cline{11-14} 
\bottomrule
\end{tabular}
}
\label{tab:different_integrated_schemes}
\end{table}
\setlength{\oldintextsep}{\intextsep}

% \vspace{-10pt}

\section{Conclusions}
This paper proposes a principled way to produce a hybrid CNNs/ViTs architecture. 
With the idea of injecting desirable inductive bias in ViTs, we present 1) a conceptual understanding of combining CNNs/ViTs into a single architecture, based on using a \textit{convolutional embedding} and its effect on the inductive bias of the architecture.
2) a conceptual framework of micro and macro detail of an hybrid architecture, where different design decisions are made at the small and large levels of detail to impose an inductive bias into the architecture.
Besides, we deliver a family of models, dubbed CETNets, which serve as a generic vision backbone and achieve the SOTA performance on various vision tasks under constrained data size.
We hope that what we found could raise a rethinking of the network design and extend the limitation of the hybrid CNNs/ViTs network.

\clearpage
\bibliographystyle{splncs04}
\bibliography{reference}
% \input{3627.bbl}
% \bibliographystyle{IEEEtran}
% \bibliography{3627}

\end{document}